\documentclass[runningheads,a4paper]{llncs}

\usepackage{amssymb}
\usepackage{graphicx}
\usepackage{multirow}
\usepackage{color}
\usepackage[colorlinks,linkcolor=blue]{hyperref}
\usepackage{epstopdf}

\usepackage{url}
\urldef{\mailsa}\path|{alfred.hofmann, ursula.barth, ingrid.haas, frank.holzwarth,|
\urldef{\mailsb}\path|anna.kramer, leonie.kunz, christine.reiss, nicole.sator,|
\urldef{\mailsc}\path|erika.siebert-cole, peter.strasser, lncs}@springer.com|    
\newcommand{\keywords}[1]{\par\addvspace\baselineskip
\noindent\keywordname\enspace\ignorespaces#1}

\begin{document}
\mainmatter  % start of an individual contribution
\title{Modeling 4D fMRI Data via Spatio-Temporal Convolutional Neural Networks (ST-CNN)}
\titlerunning{  }
\author{Yu Zhao\inst{1\dagger},Xiang Li\inst{2\dagger},Wei Zhang\inst{1},Shijie Zhao\inst{3},Milad Makkie\inst{1},\\Mo Zhang\inst{4},Quanzheng Li\inst{2\ddagger},Tianming Liu\inst{1\ddagger}}
\institute{$^{1}$Department of Computer Science, University of Georgia;\\
$^{2}$MGH/BWH Center for Clinical Data Science, Massachusetts General Hospital; \\
$^{3}$School of Automation, Northwestern Polytechnical University;\\
$^{4}$Center for Data Science, Peking University; \\
$^{5}$Laboratory for Biomedical Image Analysis, Beijing Institute of Big Data Research; \\
$\dagger$ Joint First Authors, $\ddagger$ Joint Corresponding Authors}
\authorrunning{  }
\toctitle{Lecture Notes in Computer Science}
\tocauthor{Authors' Instructions}
\maketitle
\begin{abstract}
Simultaneous modeling of the spatio-temporal variation patterns of brain functional network from 4D fMRI data has been an important yet challenging problem for the field of cognitive neuroscience and medical image analysis. Inspired by the recent success in applying deep learning for functional brain decoding and encoding, in this work we propose a spatio-temporal convolutional neural network (ST-CNN) to jointly learn the spatial and temporal patterns of targeted network from the training data and perform automatic, pin-pointing functional network identification. The proposed ST-CNN is evaluated by the task of identifying the Default Mode Network (DMN) from fMRI data. Results show that while the framework is only trained on one fMRI dataset, it has the sufficient generalizability to identify the DMN from different populations of data as well as different cognitive tasks. Further investigation into the results show that the superior performance of ST-CNN is driven by the jointly-learning scheme, which capture the intrinsic relationship between the spatial and temporal characteristic of DMN and ensures the accurate identification. 
\end{abstract}
%%%%%%%%%%%%%%%%%%%%%%%%%%%%%%%%%%%%%%%%%%%%%%%%%%%%%%%%%%%%%%%%%%%%
\section{Introduction}
%%%%%%%%%%%%%%%%%%%%%%%%%%%%%%%%%%%%%%%%%%%%%%%%%%%%%%%%%%%%%%%%%%%%
Recently, analytics of the spatio-temporal variation patterns of functional Magnetic Resonance Imaging fMRI \cite{1} has been substantially advanced through machine learning (e.g. independent component analysis (ICA) \cite{2}\cite{3} or sparse representation \cite{4}) and deep learning methods \cite{5}. As fMRI data are acquired as series of 3D brain volumes during a span of time to capture functional dynamics of the brain, the spatio-temporal relationships are intrinsically embedded in the acquired 4D data which need to be characterized and recovered. 

In literatures, the spatio-temporal analytics methods can be summarized into two groups: the first group performs the analysis on either spatial or temporal domain based on the corresponding priors, then regress out the variation patterns in the other domain. For example, temporal ICA identifies the temporally independent “signal source” in the 4D fMRI data, then obtains the spatial patterns of those sources through regression. Recently proposed deep learning-based Convolutional Auto-Encoder (CAE) model \cite{6},  temporal time series, and spatial maps are regressed later using resulting temporal features. Sparse representation methods, on the other hand, identify the spatially sparse components of the data, while the temporal dynamics of these components are obtained through regression. Works in \cite{7} utilizes Restricted Boltzmann Machine (RBM) for spatial feature analysis ignores the temporal feature.

The second group performs the analysis on spatial and temporal domain simultaneously. For example, \cite{8} applies Hidden Process Model with spatio-temporal “prototypes” to perform the spatio-temporal modeling. Another effective approach to incorporate temporal dynamics (and relationship between time frames) into the network modeling is through Recurrent Neural Network \cite{9}. Inspired by the superior performance and the better interpretability of the simultaneous spatio-temporal modeling, in this work we proposed a deep spatio-temporal convolutional neural network (ST-CNN) to model the 4D fMRI data. The goal of the model is to pinpoint the targeted functional networks (e.g., Default Mode Network DMN) directly from the 4D fMRI data. The framework is based on two simultaneous mappings: the first is the mapping between the input 3D spatial image series and the spatial pattern of the targeted network using a 3D U-Net. The second is the mapping between the regressed temporal pattern of the 3D U-Net output and the temporal dynamics of the targeted network, using a 1D CAE. Summed loss from the two mappings are back-propagated to the two networks in an integrated framework, thus achieving simultaneous modeling of the spatial and temporal domain. Experimental results show that both spatial pattern and temporal dynamics of the DMN can be extracted accurately without hyper-parameter tuning, despite remarkable cortical structural and functional variability in different individuals. Further investigation shows that the framework trained from one fMRI dataset (motor task fMRI) can be effectively applied on other datasets, indicating ST-CNN offers sufficient generalizability for the identification task. With the capability of pin-pointed network identification, ST-CNN can serve as a useful tool for cognitive or clinical neuroscience studies. Further, as the spatio-temporal variation patterns of the data are intrinsically intertwined within an integrated framework, ST-CNN can potentially offer new perspectives for modeling the brain functional architecture. 
%%%%%%%%%%%%%%%%%%%%%%%%%%%%%%%%%%%%%%%%%%%%%%%%%%%%%%%%%%%%%%%%%%%%%%%%%%
\section{Materials and Methods}
%%%%%%%%%%%%%%%%%%%%%%%%%%%%%%%%%%%%%%%%%%%%%%%%%%%%%%%%%%%%%%%%%%%%%%%%%%
ST-CNN takes 4D fMRI data as input and generates both spatial map and temporal time series of the targeted brain functional network (DMN) as output. Different from CNNs for image classifications (e.g. \cite{10}), ST-CNN consists of a spatial convolution network and a temporal convolution network, as illustrated in Fig.~\ref{fig:Picture1}(a). The targeted spatial network maps of sparse representation on fMRI data \cite{4} are used to train the spatial network of ST-CNN, while the corresponding temporal dynamics of the spatial networks are used to train the temporal networks.
%%%%%%%%%%%%%%%%%%%%%%%%%%%%%%%%%%%%%%%%%%%%%%%%%%%%%%%%%%%%%%%%%%%%%%%%%%
\subsection{Experimental data and preprocessing}
%%%%%%%%%%%%%%%%%%%%%%%%%%%%%%%%%%%%%%%%%%%%%%%%%%%%%%%%%%%%%%%%%%%%%%%%%%
We use the Human Connectome Project (HCP) Q1 and S900 release datasets \cite{11} for the experiments. Specifically, we use motor task-evoked fMRI (tfMRI) for training the ST-CNN, and test its performance using the motor and emotion tfMRI data from Q1 release and motor task tfMRI data S900 release. The preprocessing pipelines for tfMRI data include skull removal, motion correction, slice time correction, spatial smoothing, global drift removal (high-pass filtering), all implemented by FSL FEAT. 	
After preprocessing, we apply sparse representation method \cite{4} to decompose tfMRI data into functional networks on both training and testing data sets. The decomposition results consist of both the temporal dynamics (i.e. “dictionary atoms”) and spatial patterns (i.e. “sparse weights”) of the functional networks. The individual targeted DMN is then manually selected based on the spatial patterns of the resulting networks. The selection process is assisted with sorting the resulting network by their spatial overlap rate with the DMN template (from \cite{12}), measured by Jaccard similarity (i.e. overlap over union). We use the dictionary (1-D time series) of the selected network as ground-truth time series for training the CAE.
%%%%%%%%%%%%%%%%%%%%%%%%%%%%%%%%%%%%%%%%%%%%%%%%%%%%%%%%%%%%%%%%%%%%%%%%%%
\subsection{ST-CNN framework}
%%%%%%%%%%%%%%%%%%%%%%%%%%%%%%%%%%%%%%%%%%%%%%%%%%%%%%%%%%%%%%%%%%%%%%%%%%
\begin{figure}
\centering
\includegraphics[width =\textwidth]{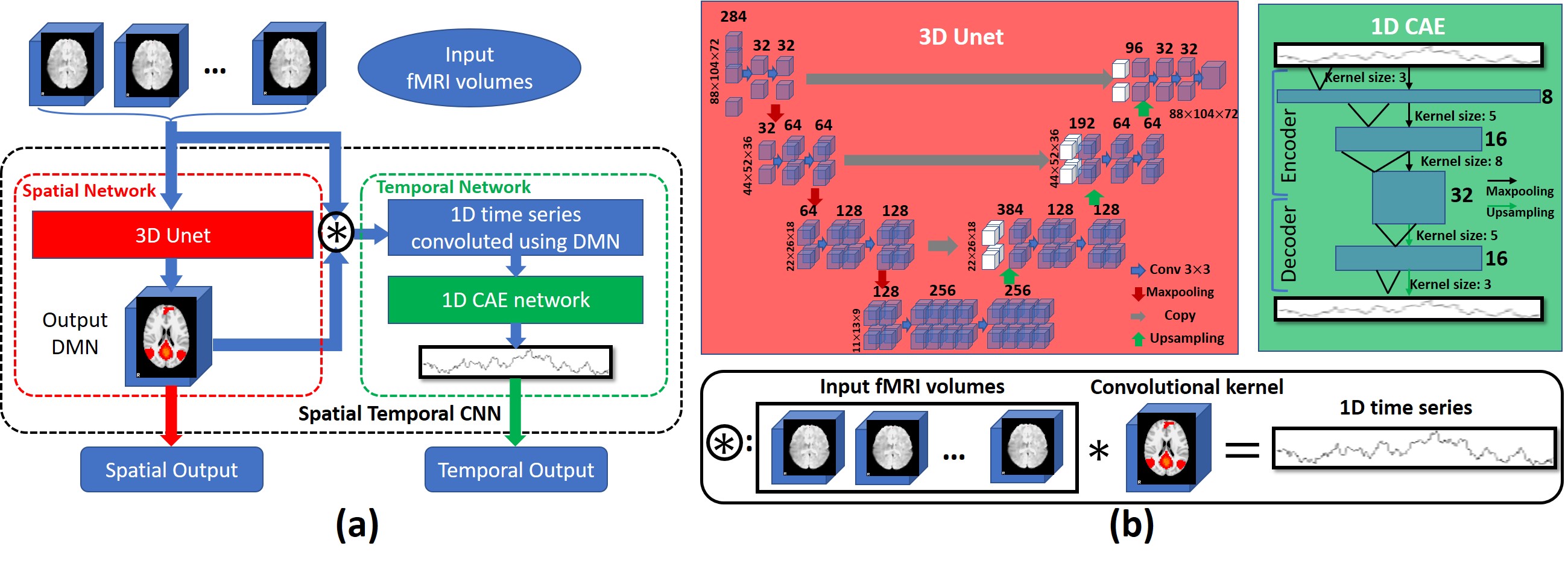}
\caption{(a). Algorithmic pipeline of ST-CNN; (b). Spatial network structure, temporal network structure, and the combination of the spatial and temporal domain.}
\label{fig:Picture1}
\end{figure}  
\textbf{Spatial Network}

The spatial network is inspired from the 2D U-Net \cite{13} for semantic image segmentation. By extending and adapting the 2D classification U-Net to a 3D regression network (Fig.~\ref{fig:Picture1}(b)), the spatial network takes 4D fMRI data as input, each 3D brain volume along the time frames is assigned to one independent channel. Basically, this 3D U-Net is constructed by a contracting CNN and a expending CNN, where the pooling layers (red arrows in Fig.~\ref{fig:Picture1}(b)) in the contracting CNN are replaced by up-sampling layers (green arrows in Fig.~\ref{fig:Picture1}(b)). This 3D U-shaped CNN structure contains only convolutional layers without fully connected layers. Loss function for training the spatial network is the mean squared error between the network output which is a 3-D image and the targeted DMN. \\
\textbf{Temporal Network}

The temporal network (Fig.~\ref{fig:Picture1}(b)) is inspired by the 1-D Convolutional Auto-Encoder (CAE) for fMRI modeling \cite{6}. Both the encoder and decoder of the 1-D CAE have the depth of 3. The encoder starts by taking 1-D signal as input and convolving it with a convolutional kernel size of 3, yielding 8 feature map channels, which are down-sampled using a pooling layer. Then a convolutional layer with kernel size 5 is attached, yielding 16 feature map channels, which are also down-sampled using a pooling layer. The last part of the encoder consists of a convolutional layer with kernel size 8, yielding 32 feature map channels. The decoder takes the output of the encoder as input and symmetrize the encoder as traditional auto-encoder structure. Loss function for training the temporal network is negative Pearson correlation (2) between the temporal CAE output time series with the temporal dynamics of the manually-selected DMN.
\begin{equation}
Temporal \ loss=-\frac{N\sum_{1}^{N}xy-\sum_{1}^{N}x\sum_{1}^{N}y}{\sqrt{(N\sum_{1}^{N}x^{2}-(\sum_{1}^{N}x)^{2})(N\sum_{1}^{N}y^{2}-(\sum_{1}^{N}y)^{2}}}
\end{equation}  \\
\textbf{Combination Joint Operator}

This combination (Fig.~\ref{fig:Picture1}(b)) procedure connects spatial network and temporal network through a convolution operator. Inputs for the combination are the 4-D fMRI data and 3-D output from the spatial network (i.e. spatial pattern of estimated DMN). The 3-D output will be used as a 3-D convolutional kernel to perform a valid no-padding convolution over each 3-D volume across each time frame of the 4-D fMRI data (3). Since the convolutional kernel size is the same as each 3D brain volume along the 4th (time) dimension, the no-padding convolution will result in a single value at each time frame, thus forming a time series for the estimated DMN. This output time series $ts$ will be used as the input for temporal 1-D CAE, as described above.
\begin{equation}
ts \in{\mathbb{R}^{T\times1}}=\{t_1,t_2,...,t_T|t_i=V_i\ast DMN \in{\mathbb{R}}\},
\end{equation}
where $t_i$ is the convolution result at each time frame, $V_i$ is the 3-D fMRI volume at time frame $i$, and DMN is the 3-D spatial network output used as convolution kernel. 
%%%%%%%%%%%%%%%%%%%%%%%%%%%%%%%%%%%%%%%%%%%%%%%%%%%%%%%%%%%%%%%%%%%%%%%%%%
\subsection{Training Process and Model Convergence}
%%%%%%%%%%%%%%%%%%%%%%%%%%%%%%%%%%%%%%%%%%%%%%%%%%%%%%%%%%%%%%%%%%%%%%%%%%
\begin{figure}
\centering
\includegraphics[width =\textwidth]{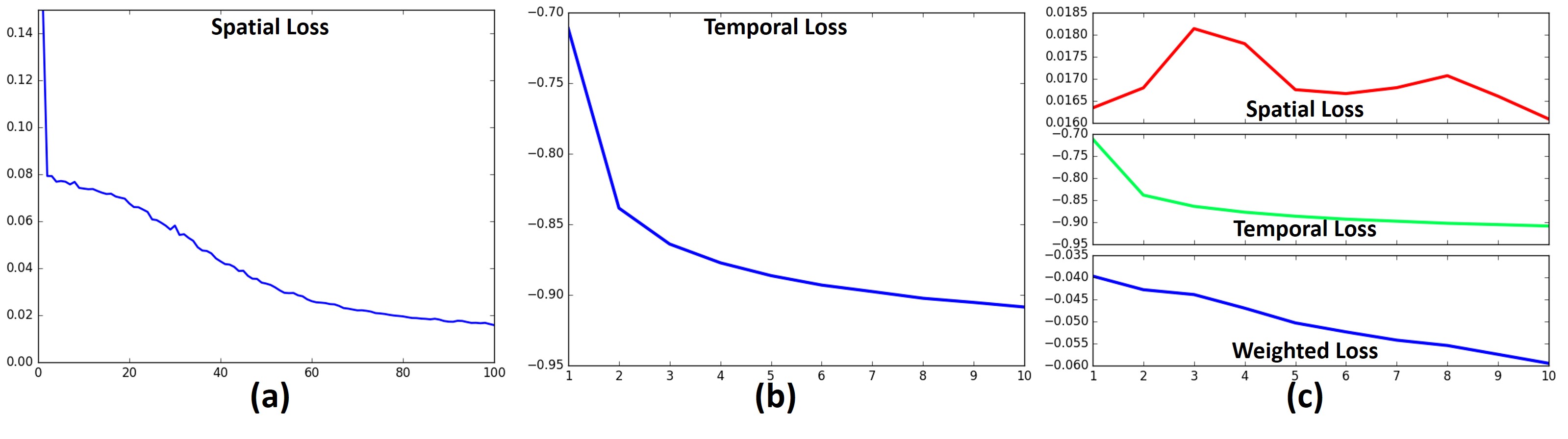}
\caption{Training losses (y-axis) versus training steps (x-axis). (a). first stage spatial network training loss; (b). second stage temporal network training loss; (c). fine-tuning training loss.}
\label{fig:Picture2}
\end{figure}
Since the temporal network will rely on the DMN spatial map from the spatial network, we split the training process into 3 stages: at the first stage, only spatial network is trained (Fig.~\ref{fig:Picture2}(a)); at the second stage, temporal network is trained based on the spatial network results (Fig.~\ref{fig:Picture2}(b)); and finally, the entire ST-CNN is trained for fine-tuning (Fig.~\ref{fig:Picture2}(c)). As we can see from Fig.~\ref{fig:Picture2}, the temporal network converges much faster (around 10 times faster) than the spatial network. Thus during the fine-tuning stage, the loss function for ST-CNN is a weighted sum (10:1) of both spatial and temporal loss.
%%%%%%%%%%%%%%%%%%%%%%%%%%%%%%%%%%%%%%%%%%%%%%%%%%%%%%%%%%%%%%%%%%%%%%%%%%
\subsection{Model Evaluation and Validation}
We firstly calculate the spatial overlap rate between the spatial pattern of ST-CNN output and a well-established DMN template to evaluate the performance of spatial network. We then calculate the Pearson correlation of the output time series with ground-truth time series from sparse representation results to evaluate the temporal network. Finally we utilize a supervised dictionary learning method \cite{14} to reconstruct the spatial patterns of the network based on temporal network result to investigate whether the spatio-temporal relationship is correctly captured by the framework.
%%%%%%%%%%%%%%%%%%%%%%%%%%%%%%%%%%%%%%%%%%%%%%%%%%%%
\section{Results}
%%%%%%%%%%%%%%%%%%%%%%%%%%%%%%%%%%%%%%%%%%%%%%%%%%%%
We use 52 subjects’ motor tfMRI data from HCP Q1 release for training the ST-CNN. We test the same trained network on three datasets: 1) motor tfMRI data from the rest of 13 subjects. 2) motor tfMRI data from 100 randomly -selected subjects in the HCP S900 release. 3) emotion tfMRI data from 67 subjects from HCP Q1 release. Testing results show consistently good performance for DMN identification, demonstrating that trained network is not limited to specific population and specific cognitive tasks.
%%%%%%%%%%%%%%%%%%%%%%%%%%%%%%%%%%%%%%%%%%%%%%%%%%%%%%%%%%%%%%%%%%%%%%%%%%
\subsection{MOTOR task testing results}
\begin{figure}
\centering
\includegraphics[width =\textwidth]{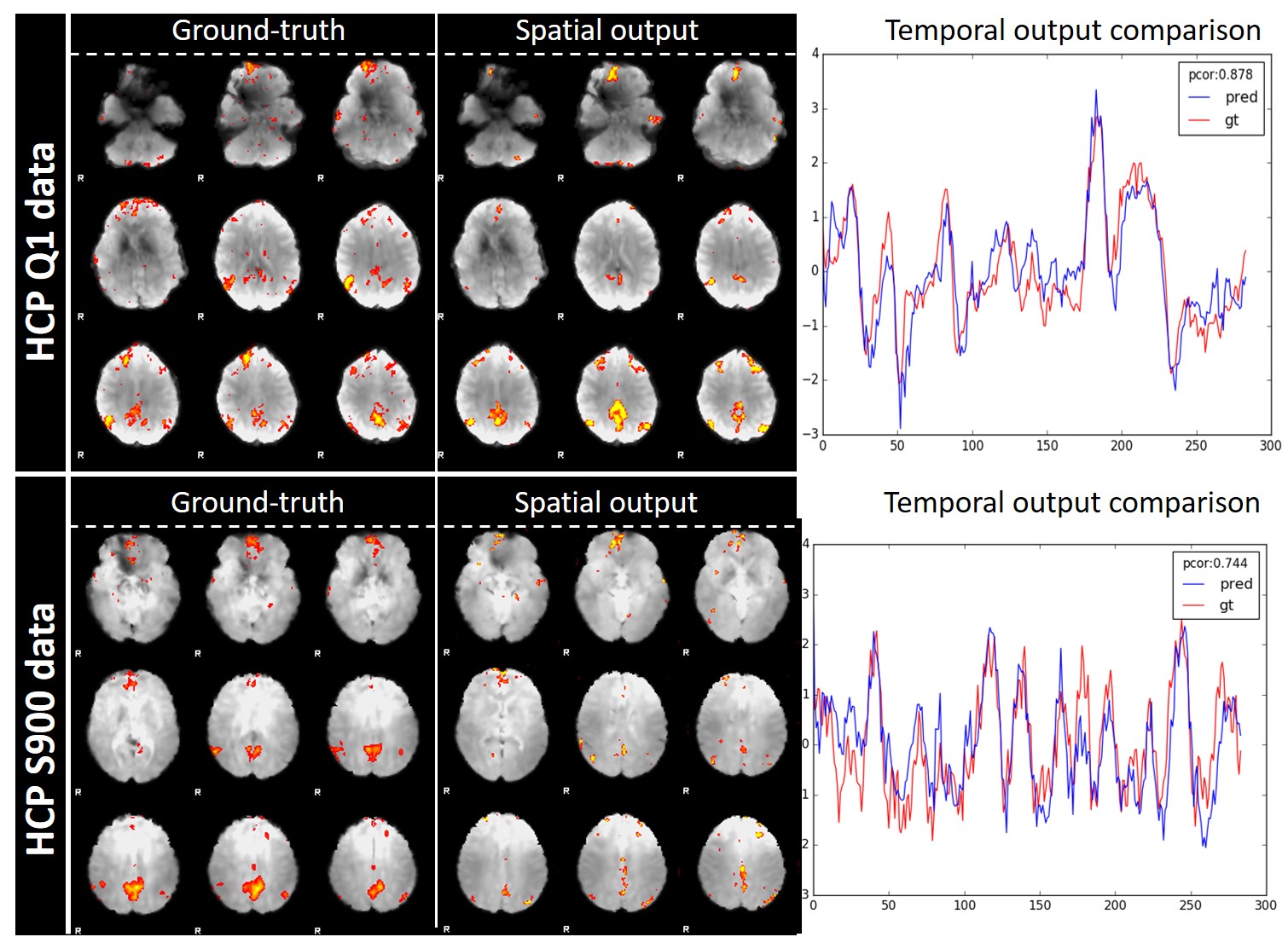}
\caption{Examples of comparisons between ST-CNN outputs and ground-truth from sparse representation. Here we showed 2 subjects’ comparison results from two different datasets (1 HCP Q1 subjects and 1 HCP S900 subjects). Spatial maps are very similar and time series have Pearson correlation coefficient values 0.878 in HCP Q1 data, and 0.744 in HCP S900 data. Red curves are ground-truth. Blue curves are ST-CNN temporal outputs.}
\label{fig:Picture3}
\end{figure}
\begin{figure}
\centering
\includegraphics[width =\textwidth]{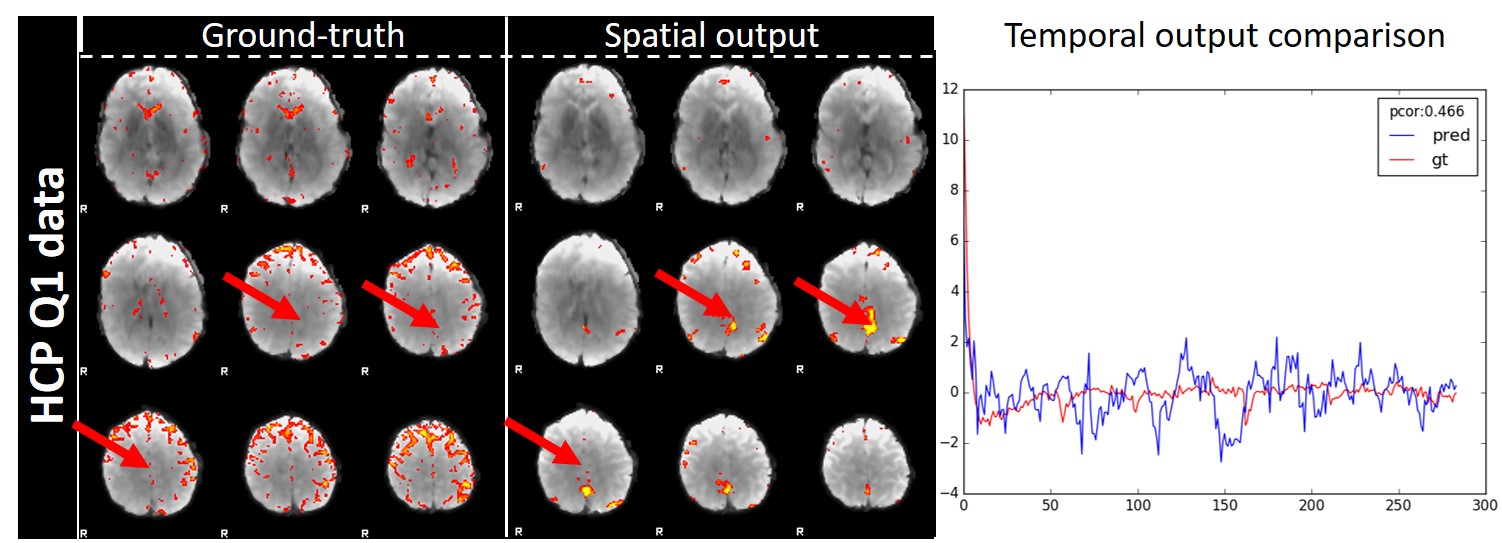}
\caption{Example of the better DMN identification of ST-CNN than sparse representation (denoted by red arrows). The temporal dynamics of the two networks are also different, where output from ST-CNN (blue) are more reasonable.}
\label{fig:Picture4}
\end{figure}

The trained ST-CNN is tested on 2 different motor task datasets: 13 subjects from HCP Q1 and 100 subjects from HCP S900, respectively. As shown in Fig.~\ref{fig:Picture3} (more examples can be found in supplemental Fig. 1), the resulting spatial and temporal patterns are consistent with the ground-truth. Quantitative analyses shown in Table~\ref{tab:table1} demonstrates that the ST-CNN performs better than sparse representation method, although it is trained from the manually-selected results of sparse representation. The rationale is that the ST-CNN can better adapt to the input data by the co-learned spatial and temporal networks, while sparse representation relies on the simple sparsity prior which can be invalid in certain cases. As shown in Fig.~\ref{fig:Picture4} (and supplemental Fig. 2), sparse representation cannot identify DMN from certain subjects while ST-CNN can. In HCP Q1 dataset, we have observed 20\% (13 out of 65 subjects) of cases where sparse representation fails while ST-CNN succeeds. Considering the fact that DMN is supposed to be consistently presented in the functioning brain regardless of task, this is an intriguing and desired characteristic of the ST-CNN model.
\begin{table}[!ht]\scriptsize
\begin{center}
\caption{Performance of ST-CNN measured by spatial overlap rate.}
\begin{tabular}{|c|c|c|c|}
\hline
\multirow{2}{*}{Datasets} 
&\multicolumn{2}{|c|}{Spatial overlap with DMN template}
& Temporal similarity \\
\cline{2-3}
& Sparse Representation &	ST-CNN & (Pearson correlation) \\
\hline
HCP Q1 MOTOR (13 subjects)  & 0.115 & \textbf{0.172} & 0.55 \\
\hline
HCP S900 MOTOR (100 subjects)  & \textbf{0.070} & 0.066 & 0.53 \\
\hline
HCP Q1 EMOTION (67 subjects)  & 0.095 & \textbf{0.168} & 0.51 \\
\hline
\end{tabular}
\label{tab:table1}
\end{center}
\end{table}
%%%%%%%%%%%%%%%%%%%%%%%%%%%%%%%%%%%%%%%%%%%%%%%%%%%%%%%%%%%%%%%%%%%%%%%%%%
\subsection{EMOTION task testing results}
\begin{figure}
\centering
\includegraphics[width =\textwidth]{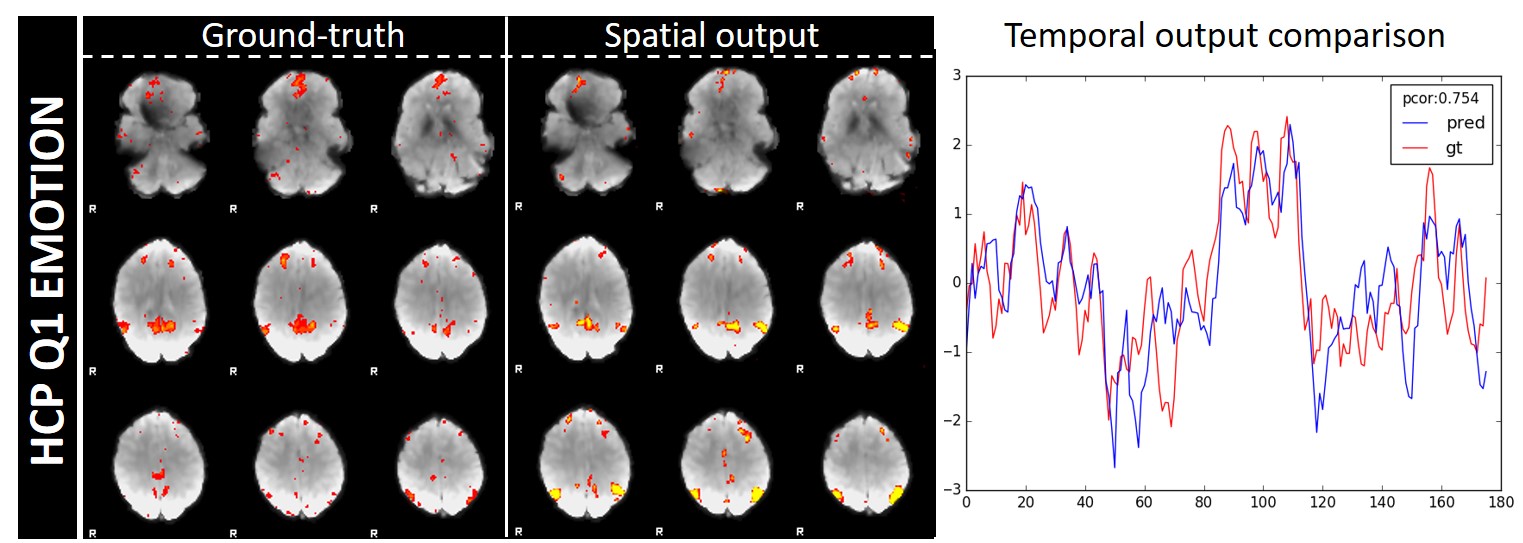}
\caption{Example of ST-CNN outputs and ground-truth (sparse representation) for EMOTION task. Spatial maps are very similar and time series have Pearson correlation 0.754. Red curve is ground-truth, blue curve is the temporal output by ST-CNN.}
\label{fig:Picture5}
\end{figure}
The 67 subjects’ emotion task-evoked fMRI data (HCP Q1) were further tested to demonstrate that our trained network based on motor task is not prone to specific cognitive tasks. The ability to extract DMN both spatially and temporally of our framework showed that the intrinsic features of DMN were well captured. As shown in Fig.~\ref{fig:Picture5} (more instances in supplemental Fig. 3), the spatial maps resemble with the ground-truth sparse representation results and so do the temporal outputs. Quantitative analyses in Table~\ref{tab:table1} showed that our outputs also had larger spatial overlap with DMN templates than outputs from sparse representation. The temporal outputs were also shown accurate, with an average Pearson correlation coefficient of 0.51. 
%%%%%%%%%%%%%%%%%%%%%%%%%%%%%%%%%%%%%%%%%%%%%%%%%%%%%%%%%%%%%%%%%%%%%%%%%%
\subsection{Spatial output and temporal output relationship}
To perform further validation, supervised sparse representation \cite{14} is applied on 13 testing subjects’ HCP Q1 motor task fMRI data. We set the temporal output of ST-CNN as predefined dictionary atoms to obtain the sparse representation on the data by learning the rest of the dictionaries. The resulting network corresponding to the predefined atom, which has the fixed temporal dynamics during the learning, are compared with ST-CNN spatial outputs. We found that the temporal output of ST-CNN can lead to an accurate estimation of the DMN spatial patterns as in Fig.~\ref{fig:Picture6} (more in supplemental Fig. 4). The average spatial overlap rate between the supervised results and ST-CNN spatial output is 0.144, suggesting that the spatial output of ST-CNN has close relationship with its temporal output.
\begin{figure}
\centering
\includegraphics[width =\textwidth]{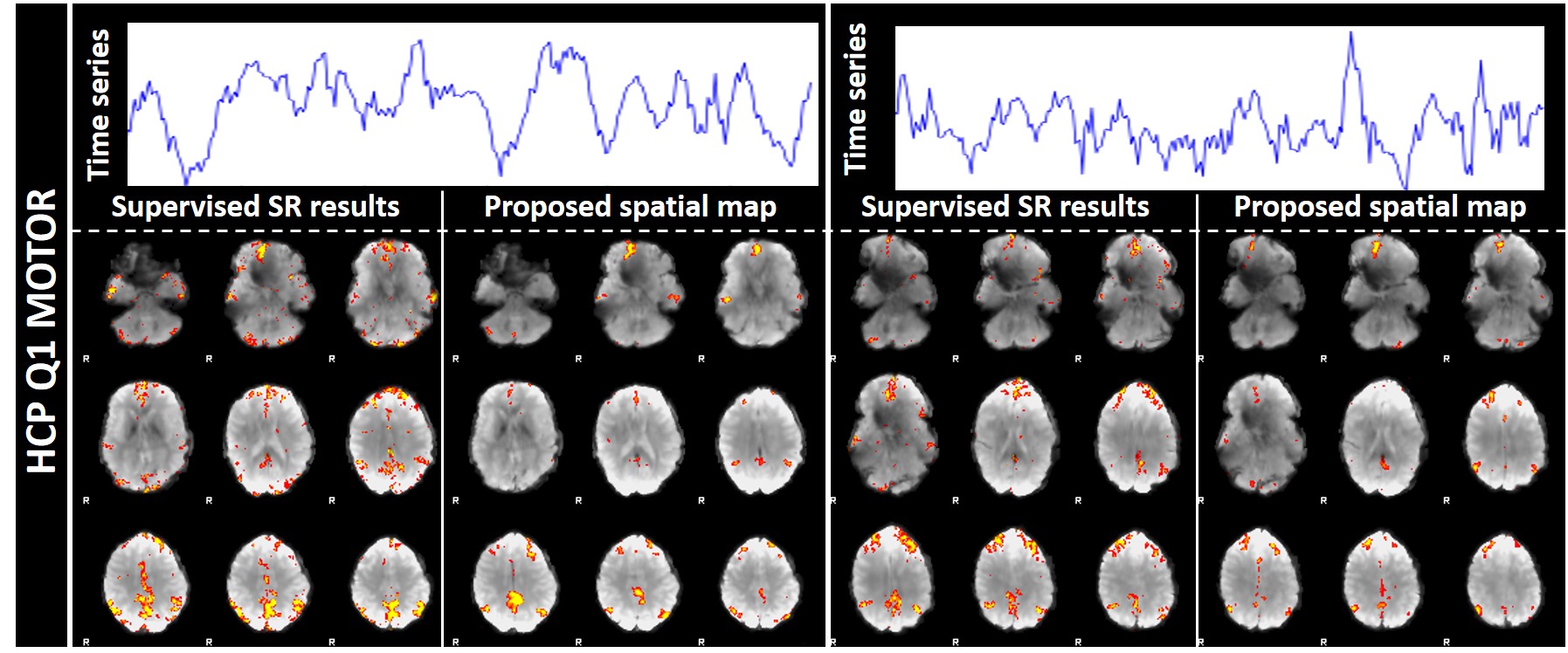}
\caption{Comparison of the spatial maps between ST-CNN results and supervised sparse representation which takes temporal output of ST-CNN as pre-defined atoms.}
\label{fig:Picture6}
\end{figure}
%%%%%%%%%%%%%%%%%%%%%%%%%%%%%%%%%%%%%%%%%%%%%%%%%%%%%%%%%%%%%%%%%%%%%%%%%%
\section{Discussion}
%%%%%%%%%%%%%%%%%%%%%%%%%%%%%%%%%%%%%%%%%%%%%%%%%%%%%%%%%%%%%%%%%%%%%%%%%%
In this work, we proposed a novel spatio-temporal CNN model to identify functional networks (DMN as an example) from 4D fMRI data modelling. The effectiveness of ST-CNN is validated by the experimental results on different testing datasets. From an algorithmic perspective, the result shows that ST-CNN embeds the spatial-temporal variation patterns of the 4D fMRI signal into the network, rather than learns the matrix decomposition process by the sparse representation. It is then very important to further refine the framework by training it over DMNs identified by other methods (such as temporal ICA). More importantly, we use DMN as a sample targeted network in the current work, due to the fact that it should be present in virtually any fMRI data. As detecting the absence/disruption a functional network is as important as identifying it (e.g. for AD/MCI early detection), in the future work we will focus on extending the current framework to pinpoint more functional networks, including task-related networks which should be presented in a limited range of datasets. We will also test ST-CNN on fMRI from abnormal brains for its capability of characterizing the spatio-temporal patterns of the disrupted DMNs.
%%%%%%%%%%%%%%%%%%%%%%%%%%%%%%%%%%%%%%%%%%%%%%%%%%%%%%%%%%%%%%%%%%%%%%%%%%
\bibliographystyle{unsrt}
\bibliography{reference.bib}

\clearpage
\newpage
%%%%%%%%%%%%%%%%%%%%%%%%%%%%%%%%%%%%%%%%%%%%%%%%%%%%%%%%%%%%%%%%%%%%%%%%%%%%%%%%%%%%%%%%%
\appendix
\renewcommand{\figurename}{Supplemental Fig.}
\setcounter{figure}{0}

\section*{\Large Supplementary Material}
\vspace{1ex}
\begin{figure}
\centering
\includegraphics[width =\textwidth]{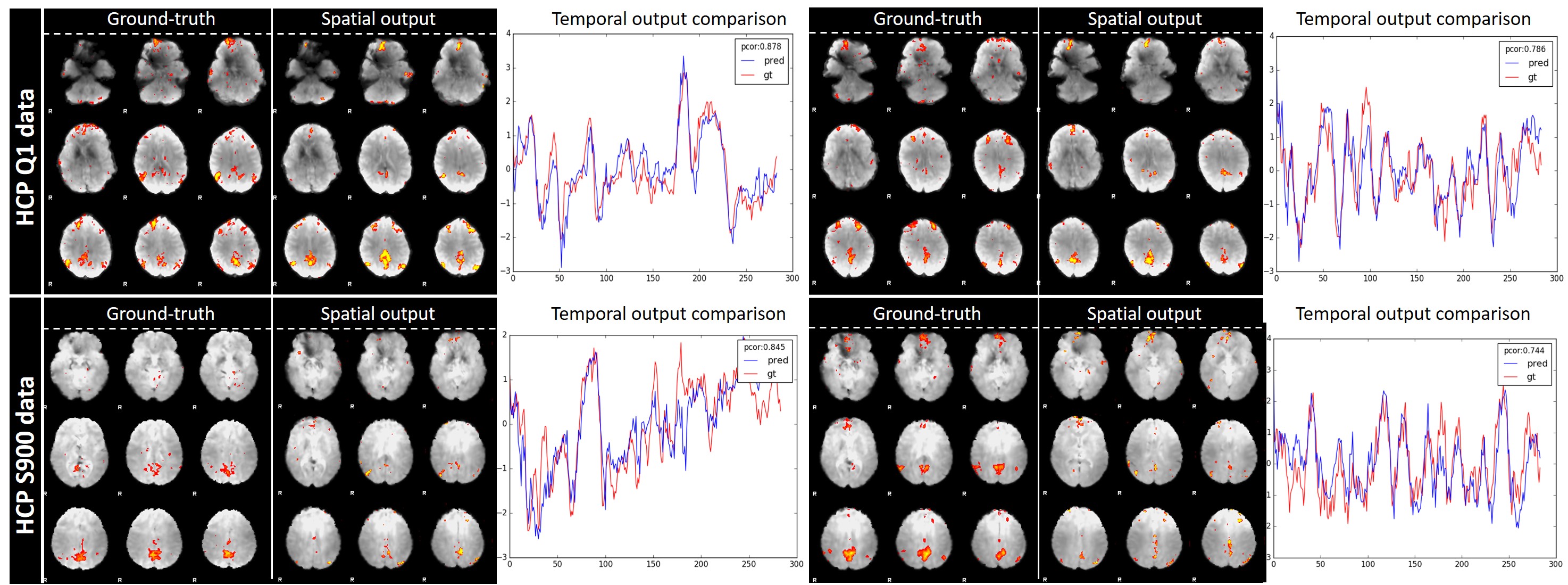}
\caption{Examples of comparisons between proposed outputs and ground-truth from sparse representa-tion. Here we showed 4 subjects’ comparison results from two different datasets (2 HCP Q1 subjects and 2 HCP S900 subjects). Spatial maps are very similar and time series have Pearson correlation coefficient values 0.878 and 0.786 in HCP Q1 data, 0.845 and 0.744 in HCP S900 data. Red curves are ground-truth. Blue curves are our proposed temporal outputs.}
\label{fig:sPicture1}
\end{figure}
\begin{figure}
\centering
\includegraphics[width =\textwidth]{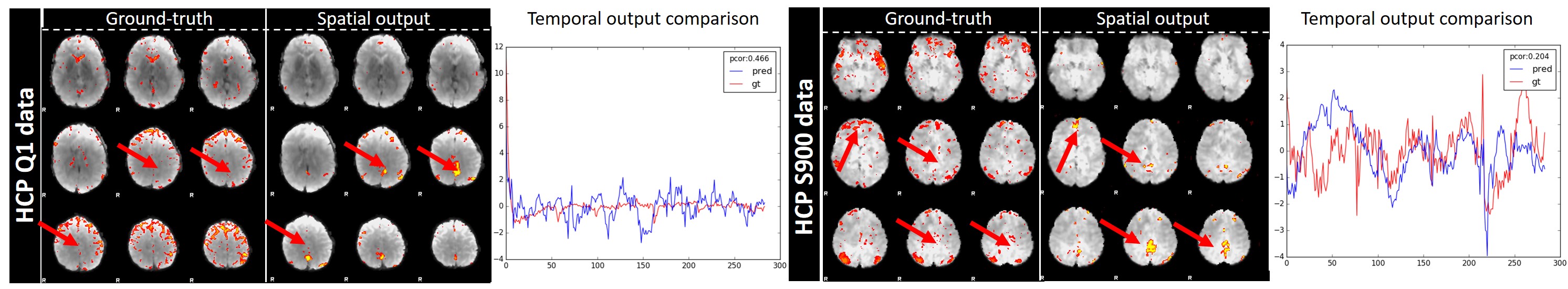}
\caption{Examples of the better DMN reconstruction ability of the proposed method than ground-truth sparse representation method (denoted by red arrows). Correspondingly, the temporal curves are not similar anymore, where our temporal output curves (blue line) are much more reasonable than noisy red line.}
\label{fig:sPicture2}
\end{figure}
\begin{figure}
\centering
\includegraphics[width =\textwidth]{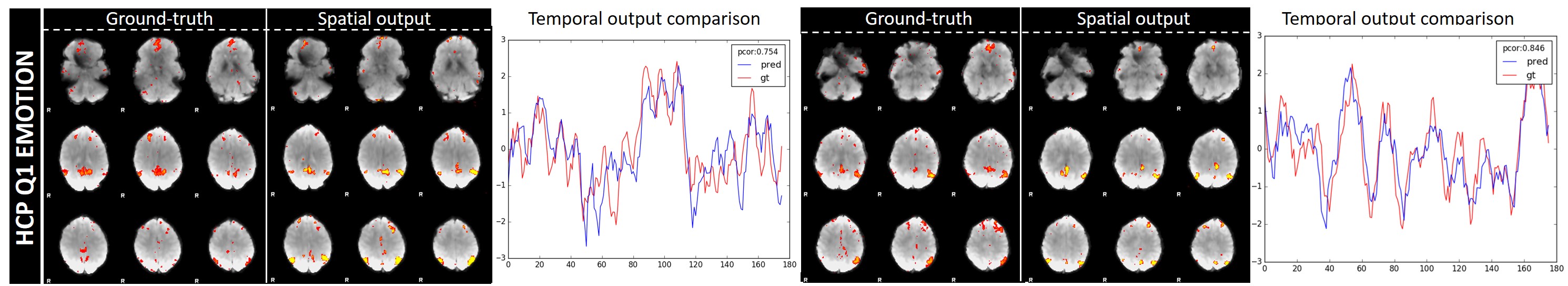}
\caption{2 examples of comparisons between proposed outputs and ground-truth from sparse representa-tion for EMOTION task. Spatial maps are very similar and time series have Pearson correlation coefficient values of 0.754 and 0.846, respectively. Red curves are ground-truth while blue curves are our proposed temporal outputs.}
\label{fig:sPicture3}
\end{figure}
\begin{figure}
\centering
\includegraphics[width =\textwidth]{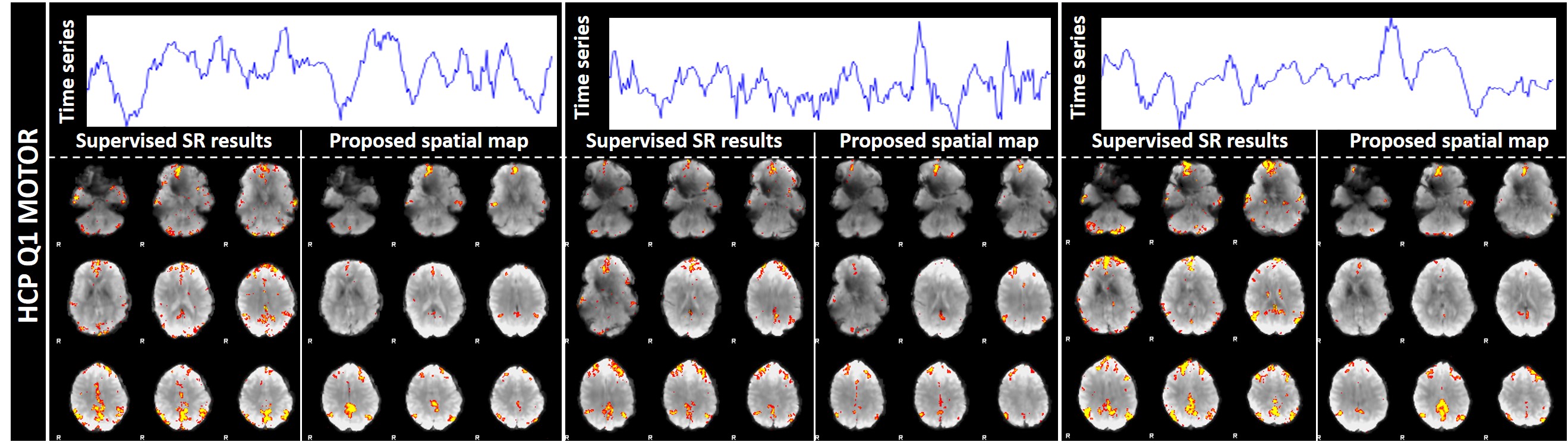}
\caption{DMN spatial map comparison between the proposed output and the reconstructed spatial map using supervised sparse representation (supervised SR) taking time series as supervision.}
\label{fig:sPicture4}
\end{figure}

\end{document}